\documentclass[conference]{IEEEtran}
\IEEEoverridecommandlockouts

\usepackage{cite}
\usepackage{amsmath,amssymb,amsfonts}
\usepackage{algorithmic}
\usepackage{graphicx}
\usepackage{subcaption}
\usepackage{textcomp}
\usepackage{xcolor}
\def\BibTeX{{\rm B\kern-.05em{\sc i\kern-.025em b}\kern-.08em
		T\kern-.1667em\lower.7ex\hbox{E}\kern-.125emX}}
\begin{document}

\title{A UAV-Mounted Sensor Network for Close-Range Inspection of Wind Turbine Rotor Blades\\
\thanks{This work was supported by the German Federal Ministry for Economic Affairs and Climate Action (BMWK) and the Projektträger Jülich (PTJ) under grant no.\ 020E-100576875 (project WISP).}
}

\author{\IEEEauthorblockN{1\textsuperscript{st} Jost Wittmann}
\IEEEauthorblockA{\textit{Department of Computer Science XVII: Robotics} \\
\textit{Julius-Maximilians Universit\"{a}t W\"{u}rzburg}\\
W\"{u}rzburg, Germany \\
Jost.Wittmann@Uni-Wuerzburg.de}
\and
\IEEEauthorblockN{2\textsuperscript{nd} Ahmed Abdullahi Hassan}
\IEEEauthorblockA{\textit{Research and Development} \\
	\textit{SubseaScanning AS}\\
	Stavanger, Norway \\
	Ahmed.Hassan@SubseaScanning.com}
\and
\IEEEauthorblockN{3\textsuperscript{rd}Nils K\"{o}mpe}
\IEEEauthorblockA{\textit{Elektronik-Entwicklung} \\
	\textit{Oktopus GmbH}\\
	Kiel, Germany \\
	NKoempe@Oktopus-Kiel.eu}
\and
\IEEEauthorblockN{4\textsuperscript{th} Prof. Dr. Andreas N\"{u}chter}
\IEEEauthorblockA{\textit{Department of Computer Science XVII: Robotics} \\
\textit{Julius-Maximilians Universit\"{a}t W\"{u}rzburg}\\
W\"{u}rzburg, Germany \\
andreas.nuechter@uni-wuerzburg.de}
}
\maketitle

\begin{abstract}
	Inspection of offshore wind turbine rotor blades is critical for predictive maintenance to maximise efficiency and extend operational lifetime. However, it remains a challenging task due to remote locations, large structural dimensions, and the limitations of current UAV-compatible sensor systems. While existing approaches can detect certain types of surface anomalies, reliable classification of defect types often remains a manual and error-prone process.
	
	This paper presents the design of a UAV-mounted multimodal sensor network combining an industrial RGB camera, a passive thermal infrared camera, and an in-house developed 3D scanner. All sensors are co-calibrated into a common coordinate frame, enabling spatial superimposition of geometric, colour, and thermal data. The system is designed to operate at close range, addressing three fundamental sensing challenges: platform motion, large field of view, and millimetre-level measurement accuracy. Preliminary laboratory results demonstrate synchronised multi-sensor acquisition and initial point cloud reconstructions, forming the basis for future airborne inspection trials.
\end{abstract}


\section{Introduction}

Maintenance of offshore wind farms is costly and hazardous. Rotor blades are susceptible to erosion, lightning strikes, and structural fatigue, yet their remote location and size make access difficult and expensive. Timely detection and accurate classification of blade anomalies are therefore critical for operational efficiency and predictive maintenance.

Successful classification requires high-quality data from multiple sensors. Close-range inspection with the turbine stopped significantly improves data quality, while standoff approaches typically lack the resolution to capture fine features such as cracks or shallow erosion pits. A UAV-mounted platform capable of close-range inspection is therefore required.

Close-range inspection of actively rotating turbines from an UAV is not feasible: blade tip speeds exceed 50\,m/s, causing turbulences and motion blur. Stopping the turbine and pitching the blades to a fixed position is thus a prerequisite for high-fidelity data acquisition \cite{b4}.

Even with stationary blades, the UAV remains a moving platform subject to wind-induced drift and vibration. Combined with the structure’s scale and defect classification accuracy requirements, this introduces three fundamental sensing challenges the proposed system must address:

\begin{itemize}
	\item \textbf{Platform motion.} The UAV moves continuously to 		
			enable time efficient inspection. In addition to this intentional motion, wind-induced drift, vibration, and gusts introduce perturbations, causing relative motion between sensor and blade that can generate motion artefacts and reduce geometric accuracy.
    \item \textbf{Large field of view.} Modern offshore rotor blades 
	       exceed 80\,m in length. Achieving full coverage at close range
	       with reasonable effort demands a wide FoV and high acquisition speed.
    \item \textbf{High measurement accuracy.} Reliable defect classification
          demands precise surface data that standard LiDAR sensors at UAV standoff distances may not provide.          
\end{itemize}

This paper presents a conceptual design for a
UAV-mounted multimodal sensor network that addresses these challenges.
By combining complementary sensing modalities and applying measures
for motion compensation, the system aims to capture the rich,
co-registered datasets required for robust anomaly classification. Prior work has demonstrated the feasibility of multimodal UAV
payloads for inspection tasks \cite{b1}, highlighted challenges
and limitations in UAV‑based visual inspection of wind turbine
blades \cite{b3}, and reported that there is no widely
established, standardised blade anomaly taxonomy in industry
practice \cite{b2}. The sensor network introduced here is
designed to address this gap.

\section{System Design}

The proposed mobile sensor network comprises three tightly integrated optical sensors. Using Zhang's calibration method \cite{b5}, each sensor produces correctly scaled datasets in a common reference frame, enabling spatial superimposition across modalities. 

Two main strategies mitigate platform motion during data acquisition. The first is minimizing acquisition time to keep motion within the accuracy threshold (\textbf{C1}). The second is motion compensation, which estimates the platform trajectory and corrects the data post-acquisition. This work focuses on the first strategy, with motion compensation as a potential future enhancement.

\subsection{Sensor Network Overview}
The sensor network is designed to be carried by a multicopter and operated at up to 
0.5\,m/s along the rotor blade at a standoff distance of 3\,m. It integrates a stereo camera pair and a laser-speckle projector to provide active 3D surface reconstruction, an RGB camera for high-resolution colour imaging, and a thermal infrared camera for temperature mapping. All sensors are mounted on a rigid frame to preserve fixed relative geometry and enable accurate co-registration of geometric, colour, and thermal data.

\begin{figure*}[t]
	\centering
	\includegraphics[width=0.8\textwidth]{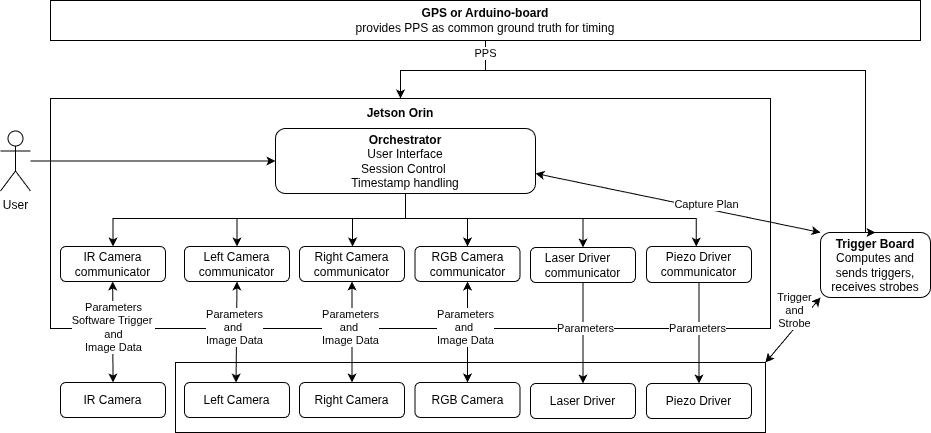}
	\caption{System architecture of the multimodal sensor network.
		All main software processes run on the central compute unit (Jetson Orin). 
		Each sensor and controller has its own communicator process. 
		The custom trigger board sends hardware trigger signals to all triggered components 
		and receives strobes in return, signaling the end of an action. 
		The thermal (IR) camera is not hardware-triggered and relies on software-based timestamp matching for synchronization.}
	\label{fig:system_overview}
\end{figure*}

\begin{figure}[t]
	\centering
	\includegraphics[width=0.7\columnwidth]{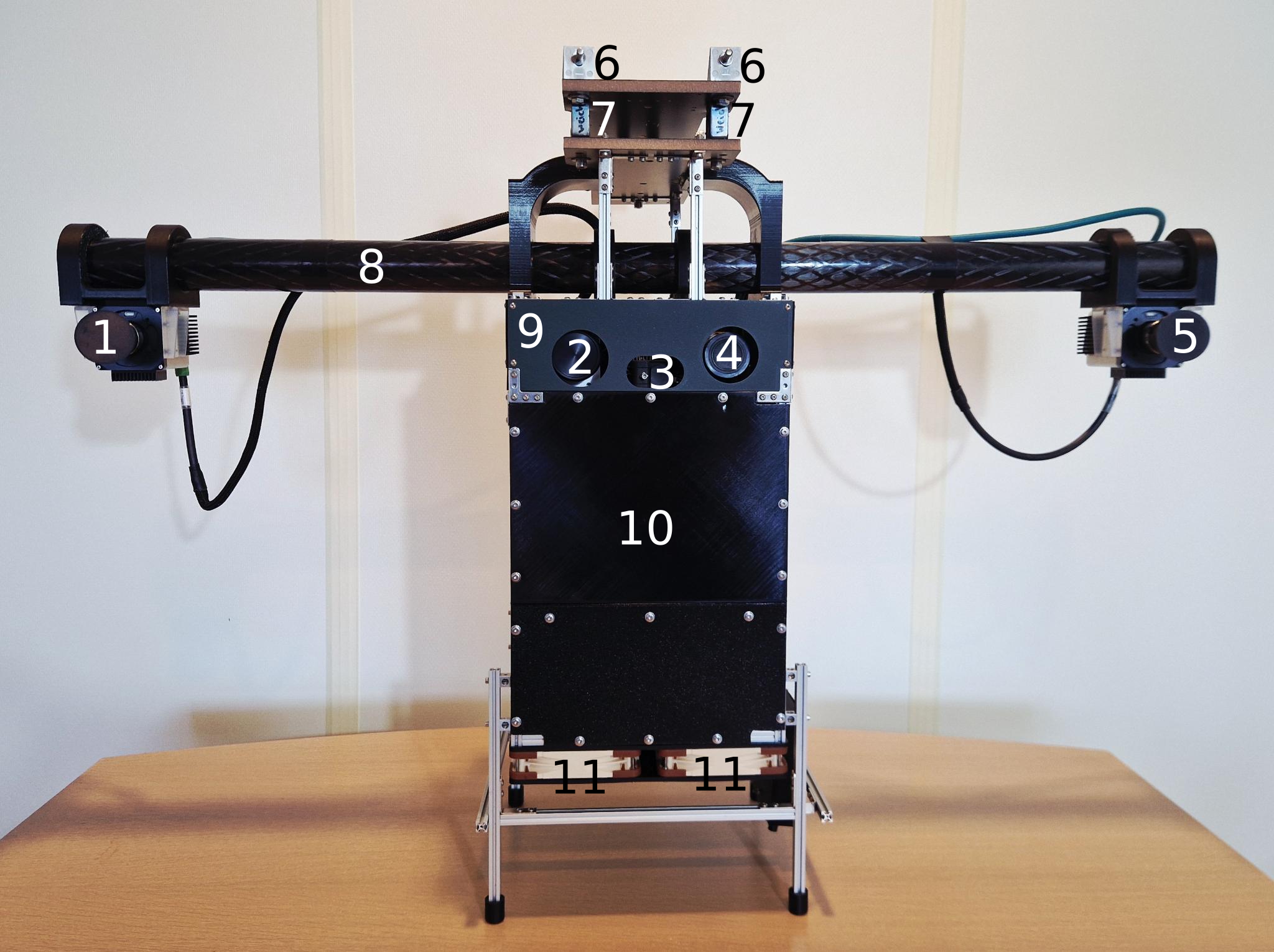}
	\caption{Prototype of the multimodal sensor payload.
		\textbf{1)}~Right stereo camera,
		\textbf{2)}~Optris PI640 thermal infrared camera,
		\textbf{3)}~Laser-speckle projector (520\,nm),
		\textbf{4)}~RGB camera,
		\textbf{5)}~Left stereo camera,
		\textbf{6)}~Mechanical attachment points to UAV,
		\textbf{7)}~Exchangeable vibration-damping mounts,
		\textbf{8)}~Carbon fibre pipe (110\,cm),
		\textbf{9)}~Rigid sensor-bearing body,
		\textbf{10)}~Non-rigid enclosure housing the Jetson Orin compute unit, controllers, and power supply,
		\textbf{11)}~2$\times$2 cooling fans (laboratory use only).}
	\label{fig:sensor_photo}
\end{figure}

The sensor payload is mounted on a rigid frame combining aluminium profiles and a carbon fibre pipe, providing a stereo baseline of approximately 1\,m while ensuring stable extrinsic calibration. Exchangeable vibration-damping mounts are integrated between the UAV interface and the sensor frame to mitigate motion-induced distortion. A lower compartment houses the onboard compute unit (Jetson Orin), power supply, and control electronics, including a custom trigger board for hardware-synchronised data acquisition and microcontrollers for projector control. The system has a total mass of 9.5\,kg and is balanced around its mounting interface.

Active laser illumination combined with a bandpass filter for the stereo-cameras increases independence of ambient lighting conditions and ensures consistent surface contrast. Direct sunlight during inspection is expected to remain a limiting factor.

The RGB camera lens is selected with a slightly larger FoV, supporting optional odometry at a later stage. 

The thermal camera comes with an uncooled bolometer sensor. Its lens approximates the FoV of the stereo system, operating in the 8–14\,µm spectral range. 

A single hardware-triggered RGB image is captured immediately following the 3D scan sequence, paired with the closest-matching thermal frame selected via software synchronization based on trigger timestamps.

\subsection{In-House Developed 3D Scanner}
The 3D scanner is the core contribution of the sensor network. It is developed
in-house to meet the specific requirements of UAV-based close-range inspection:
millimetre surface accuracy (C3), minimal acquisition time per
frame (C1), and a compact, lightweight form factor compatible with
UAV payload constraints.

\subsubsection{Measurement Principle}
The scanner employs a stereo camera pair with a baseline of approximately 1\,m.
A centrally mounted laser speckle projector provides active illumination to ensure sufficient surface texture on the typically uniform, matte surface of composite rotor blades.  
Disparity maps are computed using two complementary methods: Semi-Global Matching (SGM) \cite{b6} operates on a single stereo image pair and is applied online during acquisition. The resulting sparse point cloud is compressed and transmitted to the ground control station in real time, where the pilot may use it to perform quality control, verify coverage completeness, and maintain the correct standoff distance (C1). The second method exploits spatio-temporal correlation based on Zero-mean
Normalised Cross-Correlation (ZNCC) \cite{b7}. This method requires multiple
image pairs captured with a shifted speckle pattern between each exposure,
enabling higher-accuracy disparity estimation at the cost of increased
acquisition time. Reconstruction using ZNCC is performed during
post-processing.

\subsubsection{Projector Design}
The projector generates a laser speckle pattern by passing a collimated laser
beam through a sand-blasted diffuser glass. The optics are designed to
illuminate a circular area of approximately 1.2\,m diameter at a standoff
distance of 3\,m, providing full coverage of the stereo camera field of view
(C2).

Pattern shift is achieved using an open-loop piezoelectric actuator that
deflects the beam angle, causing a lateral displacement of the projected
speckle pattern on the target surface. The stereo cameras record a total
pattern shift of approximately 100 pixels across the full piezo displacement
range, providing sufficient sub-pattern diversity for the ZNCC-based
reconstruction. The light source is a 2\,W laser diode with a wavelength of 520\,nm. Laser brightness is controlled by a custom-built laser driver, which supports continuously adjustable output from approximately 30\% to 100\% of rated
power.

This projector design was preferred as laser speckle projectors offer an
advantageous combination of low cost, mechanical robustness, fast switching
speed, and low weight.

\subsection{Sensor Calibration}
\label{sec:calibration}
Accurate multimodal fusion requires all sensors to share a common, metrically consistent coordinate frame. Intrinsic and extrinsic calibration are performed using Zhang’s method \cite{b5} as implemented in OpenCV. We adopt a pairwise extrinsic calibration strategy, referencing all sensors to one stereo camera, thereby establishing a unified coordinate system for data fusion.

For the RGB and monochrome cameras, a planar glass checkerboard (8\,$\times$\,9 squares, 45\,mm) is used to ensure high calibration accuracy. The thermal camera is calibrated using a custom two-layer FR4 PCB featuring a 7\,$\times$\,8 checkerboard pattern with 50\,mm squares, selected to accommodate the VGA resolution of the IR camera. The differing emissivity of copper and substrate regions creates a clear thermal pattern when heated, enabling reliable corner detection in infrared imagery. See Figure ref{fig:ir-calib}

\begin{figure}[t]
	\centering
	\includegraphics[width=0.7\columnwidth]{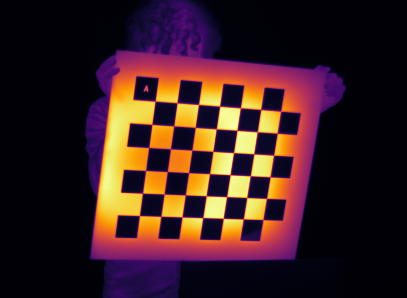}
	\caption{Custom two-layer FR4 PCB calibration board for thermal camera. Copper and substrate regions provide distinct emissivity contrast when heated, enabling reliable corner detection.}
	\vspace{-10pt}
	\label{fig:ir-calib}
\end{figure}

\subsubsection{High-Speed Data Acquisition}
Minimising acquisition time per scan frame is our primary strategy for
reducing motion-induced distortion on the UAV platform (C1). 
Table \ref{tab:acquisition} summarises the key parameters governing
acquisition speed.

\begin{table}[t]
	\caption{High-Speed Data Acquisition Parameters}
	\label{tab:acquisition}
	\centering
	\begin{tabular}{p{1.7cm} p{2.2cm} p{3.5cm}}
		\hline
		\textbf{Component} & \textbf{Parameter} & \textbf{Value} \\
		\hline
		Stereo Cameras & Sensor type & 2/3-inch, global shutter, mono. \\
		& Interface & 10\,GigE \\
		& Full resolution & 1920 $\times$ 1464\,px, 415\,fps \\
		& Binned ($2 \times 2$) & 960 $\times$ 732\,px, 791\,fps \\
		& Exposure during tests & 10\,ms \\
		\hline
		Bandpass Filter & Wavelength & 520\,nm (matched to projector) \\
		& Purpose & Ambient light rejection \\
		\hline
		Projector & Light source & 2\,W laser diode, 520\,nm \\
		& Laser driver & Custom-built (on/off in 1\,ms, preliminary) \\
		\hline
		Pattern Shift & Min.\ shift interval & $\leq 50\,\mu$s (preliminary) \\
		\hline
		Synchronisation \& Timestamps & Trigger system & Custom Arduino-based board (4\,\textmu s resolution, PPS-capable) \\
		\hline
	\end{tabular}
\end{table}

The stereo cameras, RGB camera, and projector pattern shifts are synchronised via a custom-built trigger board, which distributes hardware trigger signals to ensure temporally aligned data capture. Image timestamps, generated on the trigger board with a temporal resolution of 4,\textmu s, correspond to the centre of the exposure time; this resolution reflects the signal propagation through the electronics and does not include any computation time on the Arduino. The thermal camera does not support hardware triggering.

Each acquisition cycle consists of typically 8--16 stereo image pairs captured
for 3D reconstruction, with inter-frame intervals of approximately 50\,\textmu s
to allow for laser pattern shifts. The sequence is followed by a synchronized RGB and thermal image acquisition, during which laser illumination is disabled.

The thermal camera operates in continuous streaming mode, writing frames to an
internal buffer. Synchronisation is achieved in software by selecting the frame
closest to the corresponding trigger timestamp.

\section{Laboratory Experiments and Preliminary Validation}
The sensor network is currently under initial laboratory evaluation, with ongoing efforts to increase acquisition speed and automate the processing pipeline. A system-wide time reference is provided by an Arduino Nano generating a pulse-per-second (PPS) signal; this will be replaced by a GNSS-based PPS in future work.

The calibration achieved reprojection errors below 0.2\,pixels for the individual cameras, with the mean stereo reprojection error slightly exceeding 0.2\,pixels.

The stereo cameras operate at an exposure time of 10\,ms per image pair, ensuring stable acquisition without frame loss; a reduction to 2--3\,ms is considered feasible. The RGB camera uses an exposure time of 15\,ms, while the thermal frame is selected in software as the one closest to the RGB trigger. Each scan consists of 10 stereo image pairs and one RGB image, resulting in a total acquisition time of 117\,ms. Infrared images are captured in parallel with the RGB image but are not included in the acquisition time due to the lack of precise timing control.

\begin{figure}[t]
	\centering
	\begin{subfigure}[t]{0.9\columnwidth}
		\centering
		\includegraphics[width=0.9\linewidth]{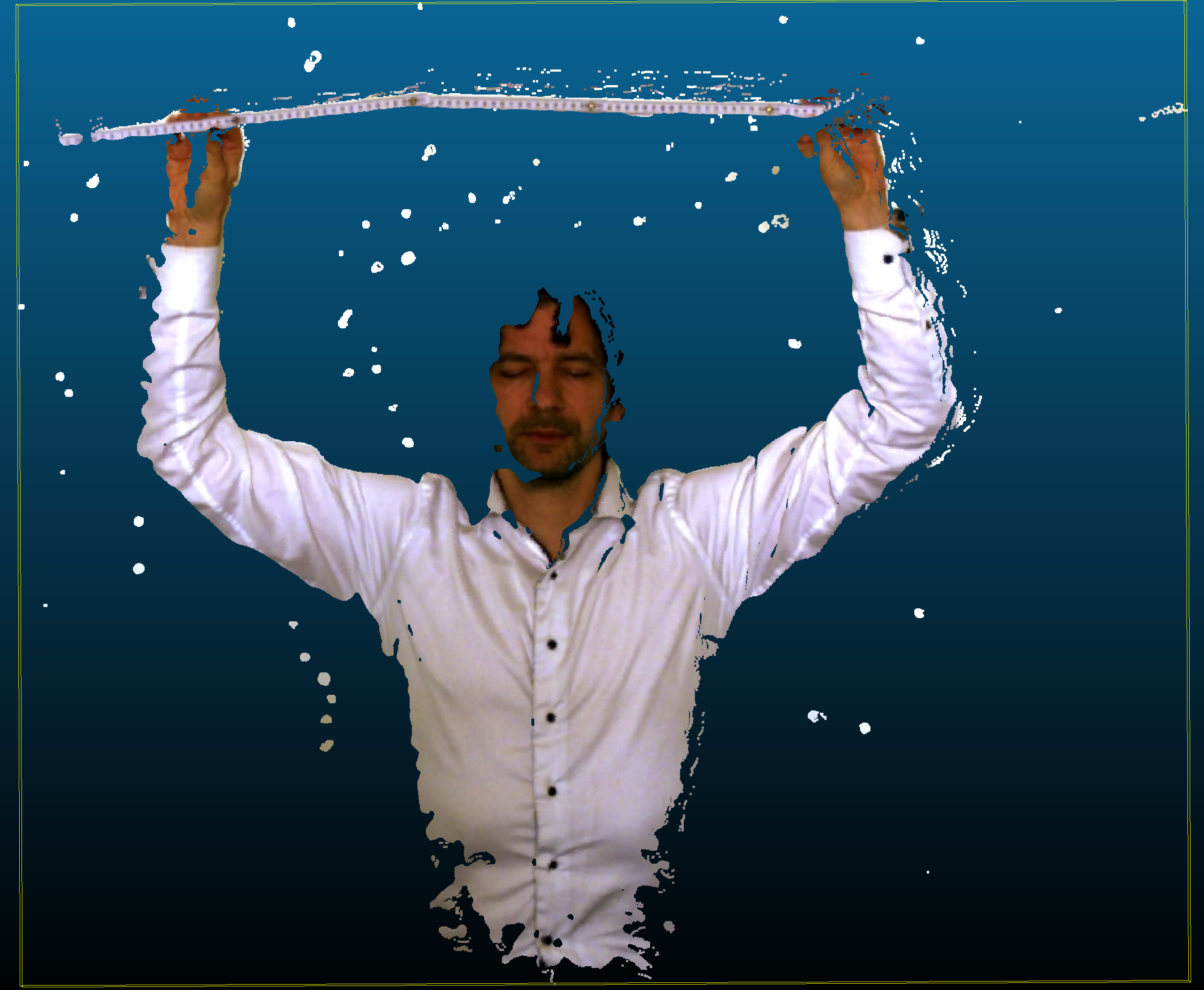}
		\caption{RGB colourisation}
		\label{fig:scan_rgb}
	\end{subfigure}
	\hfill
	\begin{subfigure}[t]{0.9\columnwidth}
		\centering
		\includegraphics[width=0.9\linewidth]{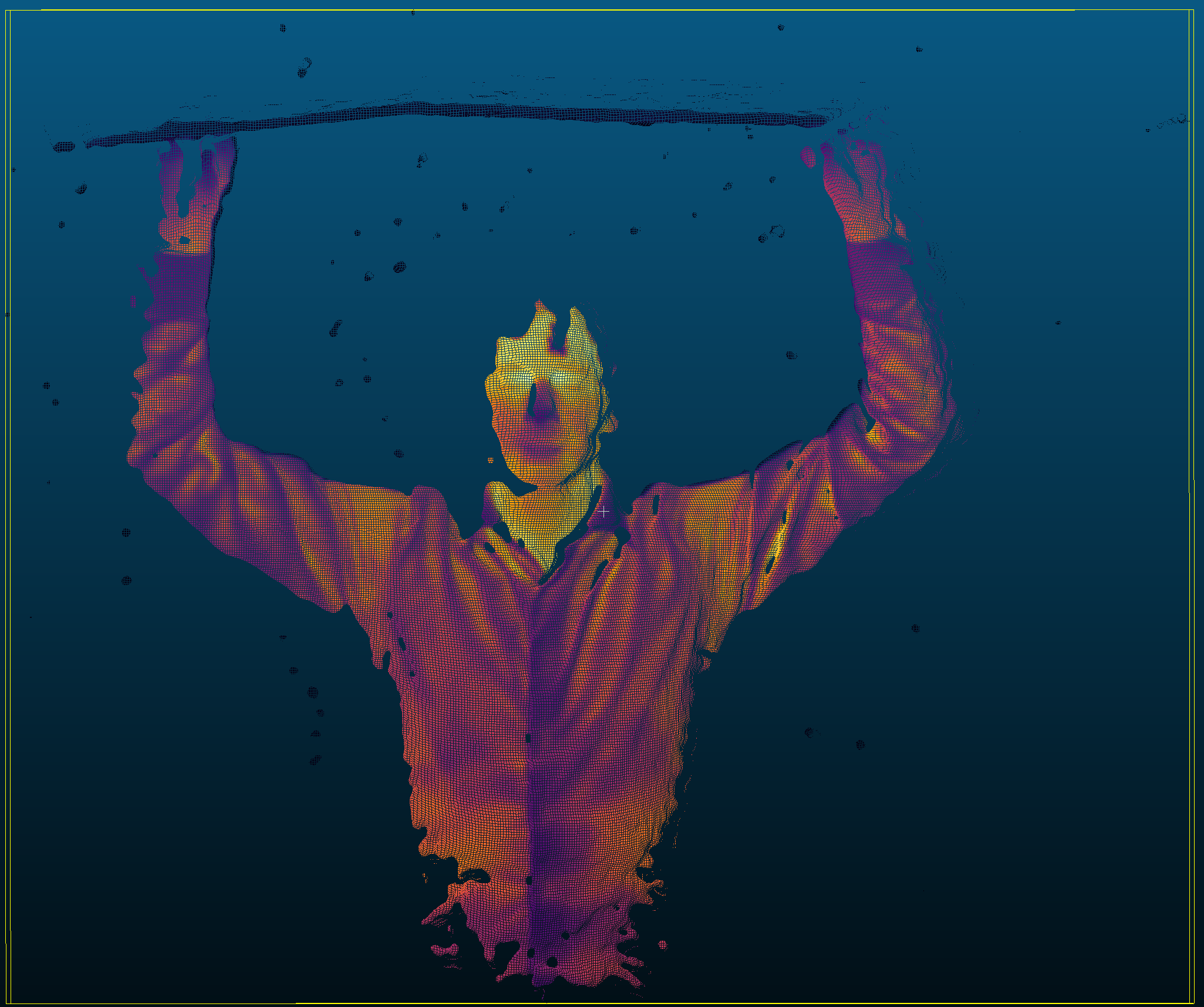}
		\caption{Thermal colourisation}
		\label{fig:scan_thermal}
	\end{subfigure}
	
	\caption{Preliminary point cloud colourisation using two modalities. The average 3D point spacing is approximately 1\,mm. The disparity maps were computed using ZNCC from 10 image pairs. At a standoff distance of 3\,m, the system captures approximately 1.3\,m of object-space width.}
	\vspace{-10pt}
	\label{fig:scan_comparison}
\end{figure}

Online disparity computation using a CUDA-based implementation of the SGM algorithm requires 0.23\,s on the Jetson Orin platform. The ZNCC algorithmen computed 406.932\,3D points shown in Fig.~\ref{fig:scan_comparison}, running for 19.8\,s per dataset on an AMD Ryzen 9 7950X CPU with an NVIDIA RTX 3090 GPU. Disparity maps were generated from 10 image pairs, followed by point cloud reconstruction and colourisation using both RGB and thermal modalities.

The reconstructed geometry demonstrates consistent surface structure across modalities. RGB colourisation provides high spatial detail, while thermal mapping enables additional inspection capabilities by highlighting temperature variations. The average 3D point spacing is approximately 1\,mm, and a scene width of 1.3\,m was captured at a 3\,m standoff distance.

\section{Conclusion and Future Work}
This paper presents the design and preliminary validation of a UAV-mounted multimodal sensor network for close-range inspection of offshore wind turbine rotor blades. The system integrates an RGB camera, thermal infrared camera, and custom 3D scanner into a co-calibrated payload that directly addresses the challenges of platform motion, wide field of view, and millimetre-scale measurement accuracy. 

These preliminary laboratory experiments confirm the feasibility of the proposed sensor network for close-range 3D reconstruction and multimodal inspection. The system demonstrates stable acquisition, accurate calibration, and reliable disparity computation under controlled conditions. Future work will focus on further reducing acquisition time, improving real-time processing, and validating system performance under UAV-based field conditions.

Future Work. The next steps include hand-held laboratory testing for motion robustness, data capture from actual rotor blades, and UAV integration for airborne trials. Additional improvements target reduced exposure times via a high-power projector design and trajectory-based motion compensation using RTK and IMU data.


\section*{Acknowledgment}
The authors thank the entire WISP project team, comprising 
Awesome Technologies Innovationslabor GmbH, Bundesanstalt f\"{u}r
Materialforschung und -pr\"{u}fung (BAM), and Oktopus GmbH, for their
contributions to the project. Further, the authors wish to thank all external contributors for making this project a success.
\end{document}